%% file: main.tex
\documentclass[nohyperref]{article}

\usepackage{microtype}
\usepackage{graphicx}
\usepackage{subfigure}
\usepackage{booktabs} 
\usepackage{multicol}
\usepackage{caption}

\usepackage{hyperref}

\usepackage[preprint]{icml2023}

\usepackage{amsmath}
\usepackage{amssymb}
\usepackage{mathtools}
\usepackage{amsthm}
\usepackage{graphicx}
\usepackage{booktabs}
\usepackage{adjustbox}
\usepackage{multirow}
\usepackage{multicol}
\usepackage{bbm}
\usepackage[capitalize]{cleveref}
\crefname{section}{Sec.}{Secs.}
\Crefname{section}{Section}{Sections}
\Crefname{table}{Table}{Tables}
\crefname{table}{Tab.}{Tabs.}

\theoremstyle{plain}

\theoremstyle{definition}

\theoremstyle{remark}

\usepackage[textsize=tiny]{todonotes}

\icmltitlerunning{Robust Camera Pose Refinement for Multi-Resolution Hash Encoding}

\begin{document}

\twocolumn[
\icmltitle{Robust Camera Pose Refinement for Multi-Resolution Hash Encoding}



\icmlsetsymbol{equal}{*}

\begin{icmlauthorlist}
\icmlauthor{Hwan Heo}{Korea,Naver}
\icmlauthor{Taekyung Kim}{Naver}
\icmlauthor{Jiyoung Lee}{Naver}
\icmlauthor{Jaewon Lee}{Korea}
\icmlauthor{Soohyun Kim}{Korea}
\icmlauthor{Hyunwoo J. Kim}{Korea}
\icmlauthor{Jin-Hwa Kim}{Naver,snu}
\end{icmlauthorlist}

\icmlaffiliation{Korea}{Department of Computer Science, Korea University, Republic of Korea}
\icmlaffiliation{Naver}{NAVER AI Lab, Republic of Korea}
\icmlaffiliation{snu}{AI Institute of Seoul National University, Republic of Korea}

\icmlcorrespondingauthor{Jin-Hwa Kim and Hyunwoo J. Kim}{j1nhwa.kim@navercorp.com; hyunwoojkim@korea.ac.kr}

\icmlkeywords{Novel View Synthesis, Neural Radiance Fields, Neural Rendering, Pose Estimation, Volume Rendering, NeRF}

\vskip 0.3in
]


\begin{NoHyper}
\printAffiliationsAndNotice{}  
\end{NoHyper}

\newcommand{\etal}{\text{et al}}
\newcommand{\eg}{\textit{e.g.}}
\newcommand{\ie}{\textit{i.e.}}
\newcommand{\rf}{\textit{ref.}}
\newcommand{\se}{$\mathfrak{se}(3) $}
\newenvironment{Figure}
  {\par\medskip\noindent\minipage{\linewidth}}
  {\endminipage\par\medskip}

\begin{abstract}
\input{0_abstract}
\end{abstract}

\input{1_Introduction}
\input{2_Related_Work}
\input{3_Methods}
\input{4_Experiments}
\input{5_Conclusion}

\newpage
\section*{Acknowledgements}
Most of this work was done while Hwan Heo and Soohyun Kim were research interns at NAVER AI Lab.
The NAVER Smart Machine Learning (NSML) platform~\cite{NSML} has been used in the experiments.

\bibliography{egbib}
\bibliographystyle{icml2023}
\input{Appendix.tex}

\end{document}

%% file: 0_abstract.tex
Multi-resolution hash encoding has recently been proposed to reduce the computational cost of neural renderings, such as NeRF.
This method requires accurate camera poses for the neural renderings of given scenes.
However, contrary to previous methods jointly optimizing camera poses and 3D scenes, the na\"ive gradient-based camera pose refinement method using multi-resolution hash encoding severely deteriorates performance.
We propose a joint optimization algorithm to calibrate the camera pose and learn a geometric representation using efficient multi-resolution hash encoding.
Showing that the oscillating gradient flows of hash encoding interfere with the registration of camera poses, our method addresses the issue by utilizing smooth interpolation weighting to stabilize the gradient oscillation for the ray samplings across hash grids.
Moreover, the curriculum training procedure helps to learn the level-wise hash encoding, further increasing the pose refinement.
Experiments on the novel-view synthesis datasets validate that our learning frameworks achieve state-of-the-art performance and rapid convergence of neural rendering, even when initial camera poses are unknown.

%% file: 1_Introduction.tex
\section{Introduction}
\label{sec:intro}
A great surge in neural rendering has emerged in the last few years.
Specifically, the Neural Radiance Fields~\cite{mildenhall2020nerf} (NeRF) has shown remarkable performances in the novel view synthesis. 
NeRF leverages a fully connected network to implicitly encode a 3D scene as a continuous signal and renders novel views through a differentiable volume rendering.
However, when the rendering of NeRF performs, a large number of inferences are inevitable, making the computational burden of training and evaluation heavier. 

Aware of this problem, related works have circumvented the shortcomings by introducing grid-based approaches~\cite{liu2020nsvf,yu2021plenoctrees,2021bakingnerf,Sun22dvgo,wu2021diver,yu2021plenoxels,2022relufields}, which store view direction-independent representations in dense grids.
While these methods explicitly encode the whole scene simultaneously, they face a trade-off between the computational cost of the model size and its performance. 
Therefore, delicate training strategies such as pruning or distillation are often required to preserve the view-synthesis quality and reduce the model size. 
Recently, Instant-NGP~\cite{mueller2022instant} addressed these problems by proposing multi-resolution hash encoding for positional encoding, which combines a multi-resolution decomposition with a lightweight hash grid. 
The multi-resolution hash encoding achieved state-of-the-art performance and the fastest convergence speed of NeRF.

\input{Figures/fig1.tex}

Despite the impressive performance of multi-resolution hash encoding, the volume rendering procedure (emission-absorption ray casting~\cite{1984raycasting}) used in Instant-NGP depends largely on the accurate camera poses. 
This method samples the points along the ray defined by direction and origin, which are determined by the camera pose.
However, obtaining accurate camera poses in real-world scenarios might be unavailable, 
so most existing works utilize an off-the-shelf algorithm such as Structure-from-Motion (SfM), or COLMAP~\cite{2016sfm}.
The previous works~\cite{wang2021nerfmm,SCNeRF2021,lin2021barf} have attempted to resolve this issue by jointly optimizing camera poses and scene representations with the original NeRF. 
However, applying this approach to multi-resolution hash encoding leads to severe deteriorations in pose refinement and scene representation.

Based on the gradient analysis of the na\"ive joint optimization of pose parameters and multi-resolution hash encodings, 
we demonstrate that the non-differentiability of the hash function and the discontinuity of the $d$-linear weights as a function of the input coordinate leads to the fluctuation in the Jacobian of the multi-resolution hash encodings. 
We investigate a novel learning strategy for jointly optimizing the camera pose parameters and the other parameters when the camera poses are noisy or unknown, utilizing the outstanding performance of multi-resolution hash encoding.

Given that, we propose to use a non-linear activation function in our straight-through estimator for smooth gradients in the backward pass, consistently maintaining the $d$-linear interpolation in the forward pass (\rf ~Figure~\ref{fig:fig1}).
Moreover, we propose the multi-level learning rate scheduling that regulates the convergence speed of each level-wise encoding.
We also empirically show that a small decoder compared to the size of the hash table~\cite{mueller2022instant} converges to suboptimal when the camera poses are noisy.
The ablation studies on the depth and wide of the decoding networks and the core components of the proposed learning framework firmly validate our proposed method for robust camera pose refinement for multi-resolution hash encoding.

In summary, our contributions are three-fold:
\begin{itemize}
    \item[\textbullet] We analyze the derivative of the multi-resolution hash encoding, and empirically show that the gradient fluctuation negatively affects the pose refinement.
    \item[\textbullet] We propose an efficient learning strategy jointly optimizing multi-resolution hash encoding and camera poses, leveraging the smooth gradient and the curriculum learning for coarse-to-fine adaptive convergences.
    \item[\textbullet] Our method achieves state-of-the-art performance in pose refinement and novel-view synthesis with  a faster learning speed than competitive methods.
\end{itemize}

%% file: Figures/fig1.tex
\begin{figure}[t]
    \vskip 0.2in
    \centering
    \includegraphics[width=\columnwidth]{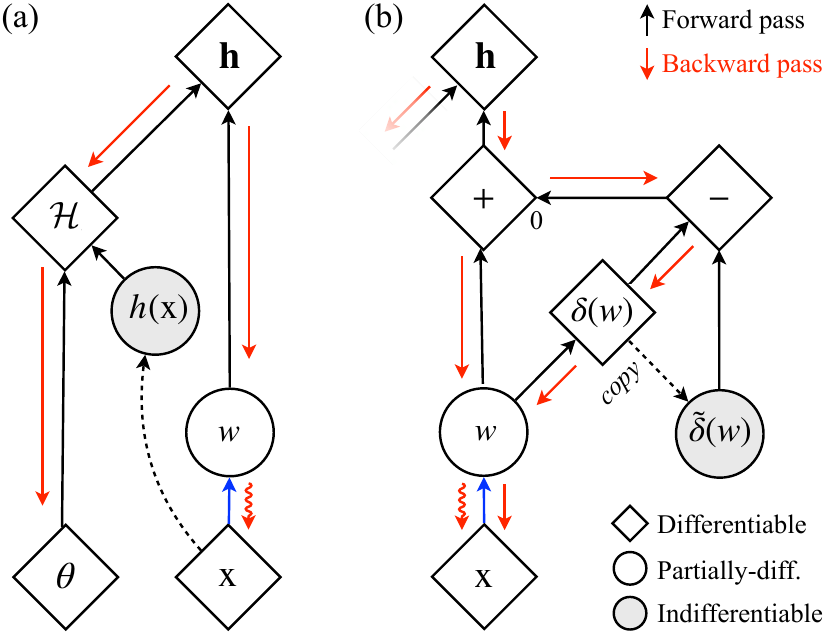}
    \vskip -0.03in  
    \caption{Gradient smoothing (a) $\rightarrow$ (b) to attenuate the gradient fluctuation (jiggled red arrow) of the hash encoding $\mathbf{h}$. For the camera pose refinement, the error back-propagation passes through the $d$-linear interpolation weight $w$; however, its derivative is determined by the sign of relative position of the input coordinate $\mathbf{x}$ to the corners of the hash grid. 
    The gradient fluctuation from this makes it difficult to converge.
    Please refer to \cref{sec:method} for the implementation details and the definitions of other symbols.
    }
    \label{fig:fig1}
    \vskip -0.2in
\end{figure}

%% file: 2_Related_Work.tex
\section{Related Work}
\label{sec:2}
\subsection{Neural Rendering}
\label{subsec:2.1} 
\citet{mildenhall2020nerf} first introduced the Neural Radiance Fields (NeRF) which parameterizes 3D scenes using neural networks. 
They employed a fully differentiable volume rendering procedure and a sinusoidal encoding to reconstruct high-fidelity details of the scene representations. 
The necessity of sinusoidal encoding was examined from the perspectives of kernel regression~\cite{tancik2020fourier}, or the hierarchical structure of a natural scene reconstruction task~\cite{2022pins}. 

Subsequently, in order to improve the reconstruction quality of NeRF, various modifications have been proposed such as replacing the ray casting with anti-aliased cone tracing~\cite{barron2021mipnerf}, disentangling foreground and background models through non-linear sampling algorithms~\cite{2020nerfpp,neff2021donerf,barron2022mipnerf360}, or learning implicit surface instead of the volume density field, \eg, signed distance function~\cite{Oechsle2021unisurf,wang2021neus,yariv2021volume}.
Also, there are several applicative studies with decomposition of NeRF~\cite{pumarola2020dnerf,martinbrualla2020nerfw,Pratul2021nerv,boss2021nerd,rebain2021derf,park2021nerfies}, composition with generative works~\cite{2021GRAF,Niemeyer2020GIRAFFE,2022clipnerf,jain2021dreamfields}, or few-shot learning~\cite{yu2021pixelnerf,jain2021dietnerf,rebain2022lolnerf,2022sinnerf,chibane2021SRF,wei2021nerfingmvs,chen2021mvsnerf}.



\subsection{Accelerating NeRF}
\label{subsec:2.2.}
One crucial drawback of NeRF is its slow convergence and rendering speed. 
To accelerate the training speed of NeRF, previous works have combined grid-based approaches which store view-direction-independent information on voxel grids. 
\citet{liu2020nsvf} introduce a dense feature grid to reduce the computation burden of NeRF and progressively prunes the dense grids.
The other works pre-compute and store a trained NeRF to the voxel grid, increasing rendering speed~\cite{yu2021plenoctrees,2021bakingnerf}.
On the other hand, rather than distilling the trained NeRF to voxel grids, direct learning of features on the voxel has been proposed~\cite{Sun22dvgo,wu2021diver,2022fplen, yu2021plenoxels}.

While these methods have been successful in achieving near real-time neural rendering, they also come with drawbacks such as the increased model size and lower reconstruction quality caused by pre-storing the scene representation.
To overcome these limitations, \citet{mueller2022instant} recently proposed Instant-NGP, which utilizes spatial hash functions and multi-resolution grids to approximate dense grid features and maximizes the hierarchical properties of 3D scenes.
This approach allows for state-of-the-art performance and the fastest convergence speed simultaneously. 

\subsection{NeRF with Pose Refinement}
\label{subsec:2.3.}
For the majority of neural rendering, it is crucial to have accurate camera intrinsic and extrinsic parameters. 
In an effort to address this issue, \citet{2020inerf} proposed a method for combining pose estimation and NeRFs by utilizing an inverted trained NeRF as an image-to-camera pose model. 
Subsequently, various methods for jointly optimizing camera pose parameters and 3D scene reconstruction have been proposed.
\citet{wang2021nerfmm} proposed a joint optimization problem in which the camera pose is represented as a 6-degree-of-freedom (DoF) matrix and optimized using a photometric loss. 
Building upon this, \citet{2022sinerf} proposed a method that replaces ReLU-based multi-layer perceptrons (MLPs) with sine-based MLPs and employs an efficient ray batch sampling.

In addition to directly optimizing camera parameters, geometric-based approaches~\cite{SCNeRF2021, lin2021barf,2022garf} have also been suggested.
For example, \citet{lin2021barf} proposed BARF, which optimizes the warping matrix of the camera pose with a standard error back-propagation algorithm, utilizing curriculum training to adjust the spectral bias of the scene representation.

Unlike previous methods based on the original NeRF structure, our method is designed for grid-based approaches, especially for multi-resolution hash encoding, which shows outstanding performance in novel-view synthesis and its training speed. 
The common NeRF structure and its variants are prone to slowly converge, but our method can be converged significantly faster with state-of-the-art reconstruction performance under the circumstance of noisy or unknown camera poses.

%% file: 3_Methods.tex
\section{Method}
\label{sec:method}
As mentioned in Section~\ref{sec:intro}, 
we observed that a na\"ive error back-propagation for the camera pose refinement with multi-resolution hash encoding leads to inferior results compared to the use of sinusoidal encoding, \citep[\eg,][]{SCNeRF2021,lin2021barf}.
To further understand the observation, we analyze the derivative of the multi-resolution hash encoding (Section~\ref{sec:3.1.}). 
We point out that the gradient fluctuation of the multi-resolution hash encoding makes it difficult to learn the pose refinement and scene reconstruction jointly (Section~\ref{sec:3.2.}). 
To address these, we propose a method for calibrating inaccurate camera poses in multi-resolution hash encoding (Section~\ref{sec:3.3.}).  
Additionally, we find that the multi-level decomposition of a scene induces the different convergence rates of multi-level encoding, which results in limited camera pose registration (Section~\ref{sec:3.4.}). 

\input{3_1_hash_encoding}
\input{3_2_pose_refine}
\input{3_3_Smooth}

%% file: 3_1_hash_encoding.tex
\subsection{Multi-Resolution Hash Encoding}
\label{sec:3.1.}
This section describes the multi-resolution hash encoding presented by \citet{mueller2022instant}, which we focus on. 

\subsubsection{Multi-Resolution Hash Encoding}
As the combination of the multi-resolution decomposition and the grid-based approach with hashing mechanism, multi-resolution hash encoding is defined as a learnable mapping of input coordinate $\mathbf{x} \in \mathbb{R}^{d}$ to a higher dimension.  
The trainable encoding is learned as multi-level feature tables independent of each other.

The feature tables $\mathcal{H} = \left \{ \mathcal{H}_{l} \ | \ l \in \left \{ 1, \dots, L \right \} \right \}$ are assigned to the $L$ levels and each table contains $T$ trainable feature vectors with the dimensionality of $F$.
Each level consists of the $d$-dimensional grids where each dimension has $N_{l}$ sizes considering multi-resolution. 
The number of grids for each size exponentially grows from the coarsest $N_{\text{min}}$ to the finest resolutions $N_{\text{max}}$. 
Therefore, $N_l$ is defined as follows:
\begin{align}
\label{eq:grid}
        b 
        \vcentcolon &= 
        \exp 
        \left( 
            \frac{\ln N_{\text{max}} - \ln N_{\text{min}}}{L-1}  
        \right) 
        \\
        N_{l} 
        \vcentcolon &= 
        \lfloor N_{\text{min}} \cdot b^{l-1} \rfloor. 
\end{align}

For given a specific level $l$, an input coordinate $\mathbf{x}$ is scaled by $N_{l}$ and then a grid spans to a unit hypercube where $\lfloor \mathbf{x}_{l} \rfloor \vcentcolon = \lfloor \mathbf{x}_{l} \cdot N_{l} \rfloor$ and $\lceil \mathbf{x}_{l} \rceil \vcentcolon = \lceil \mathbf{x}_{l} \cdot N_{l} \rceil$ are the vertices of the diagonal. 
Then, each vertex is mapped into an entry in the level’s respective feature table.
Notice that, for coarse levels where the number of the total vertices of the grid is fewer than $T$, 
each vertex corresponds one-to-one to the table entry. 
Otherwise, each vertex corresponds to the element of the $l^{\text{th}}$ table $\mathcal{H}_{l}$, whose table index is the output of the following spatial hash function~\cite{2003hash}:
\begin{equation}
\label{eq:spatial_hashing}
    h(x) 
    = 
    \left ( 
        \bigoplus_{i=1}^{d} x_{i} \pi_{i}  
    \right ) 
    \quad \text{mod } T,
\end{equation}
where $\bigoplus$ denotes bitwise XOR and $\pi_{i}$ are unique and large prime numbers. 
In each level, the $2^d$ feature vectors of the hypercube are $d$-linearly interpolated according to the relative position of $\mathbf{x}$. 
However, the interpolation enables us to get the gradient of the table entry since the interpolating weights are the function of $\mathbf{x}$. 
We will revisit this in the following section for analysis.

The output $\mathbf{y}$ of the multi-resolution hash encoding is the concatenation of the entire level-wise interpolated features and its dimensionality is $L \times F$. 
For simplicity, we denote as $\mathbf{y} = f(\mathbf{x}; \theta)$ with its trainable parameter $\theta$. 
Similar to the other neural renderings using differentiable volume rendering (emission-absorption ray casting), the decoding MLP $m(\mathbf y; \phi)$ predicts the density and the non-Lambertian color along the ray. 
All trainable parameters are updated via photometric loss $\mathcal{L}$ between the rendered ray and the ground-truth color.

\subsubsection{Derivative of Multi-Resolution Hash Encoding}
For the gradient analysis, we derive the derivative of the multi-resolution hash encoding with respect to $\mathbf{x}$. 
Let $\mathbf{c}_{i,l}(\mathbf{x})$ denote the corner $i$ of the level $l$ resolution grid in which $\mathbf{x}_{l}$ is located, and let $h_{l}(\cdot)$ represent the hash function for the $l$-level, as defined in \cref{eq:spatial_hashing}. 

Next, consider a function $\mathbf{h}_{l}: \mathbb{R}^{d} \rightarrow \mathbb{R}^{F}$, whose output is a $l^{\text{th}}$ interpolated feature vector with $2^{d}$ corners, as given by,
\begin{equation}
\label{eq:trilinear}
    \begin{split}
        \mathbf{h}_{l} (\mathbf{x})
        = 
        \sum_{i=1}^{2^{d}} w_{i, l} 
        \cdot 
        \mathcal{H}_{l}\big(h_{l}\big(\mathbf{c}_{i,l}(\mathbf{x}) \big) \big),
    \end{split}
\end{equation}
where $w_{i,l}$ denotes the $d$-linear weight, which is defined by the \textit{opposite} volume in a unit hypercube with the relative position of $\mathbf{x}$:
\begin{equation}
\label{eq:trilinear_weight}
    \begin{split}
        w_{i,l} 
        = 
        \prod_{j=1}^{d} 
        \left( 
            1 - | \mathbf{x}_{l} - \mathbf{c}_{i,l}(\mathbf{x}) |_j 
        \right )
    \end{split}
\end{equation}
where the index $j$ indicates the $j$-th dimension in the vector.
We can redefine the multi-resolution hash encoding vector $\mathbf{y}$ as follows:
\begin{equation}
\label{eq:hash_enc}
    \begin{split}
        \mathbf{y} 
        = 
        f(\mathbf{x}; \theta) 
        = 
        \big[ 
            \mathbf{h}_1(\mathbf{x}); \dots; \mathbf{h}_L(\mathbf{x}) 
        \big] 
        \in \mathbb{R}^{F'}
    \end{split}
\end{equation}
where the dimension of the output vector is $F' = L \times F$ after the concatenation.

The Jacobian $\nabla_{\mathbf{x}}\mathbf{h}_{l}(\mathbf{x}) \in \mathbb{R}^{F \times d}$ of the $l^{\text{th}}$ interpolated feature vector $\mathbf{h}_l (\mathbf{x})$ with respect to the $\mathbf{x}$ can be derived using the chain-rule as follows: 
\begin{equation}
\label{eq:grad}
    \begin{split}
        \nabla_{\mathbf{x}}\mathbf{h}_{l}(\mathbf{x}) 
        &= 
        \left[  
            \frac{\partial {\mathbf{h}_{l}}(\mathbf{x})}{\partial {x}_1}, 
            \dots,
            \frac{\partial {\mathbf{h}_{l}}(\mathbf{x})}{\partial {x}_d}
        \right]
        \\
        &= 
        \sum_{i=1}^{2^{d}} 
        \mathcal{H}_{l}\big( h_{l} \big( \mathbf{c}_{i,l} (\mathbf{x}) \big) \big)
        \cdot
        \nabla_{\mathbf{x}}{w_{i,l}},
    \end{split}
\end{equation}
where $\mathcal{H}_{l}(g_{l}(\mathbf{c}_{l}(\mathbf{x}) ))$ is not differentiable with respect to $\mathbf{x}$ and the $k$-th element of $\nabla_{\mathbf{x}}{w_{i,l}} \in \mathbb{R}^{1 \times d}$ is defined by, 
\begin{equation}
\label{eq:weight_grad}
    \begin{split}
        \frac{\partial {{w}_{i,l}}(\mathbf{x})}{\partial {x}_k} 
        &= 
        s_k
        \cdot 
        \prod_{j \neq k} 
        \left(
            1 - | \mathbf{x}_{l} - \mathbf{c}_{i,l}(\mathbf{x}) |_j 
        \right ),
    \end{split}
\end{equation}
where $s_k$ denotes $s_k = \textit{sign}\big(|\mathbf{c}_{i,l}(\mathbf{x}) - \mathbf{x}_{l}|_k \big)$.

As seen in \cref{eq:grad}, the Jacobian $\nabla_{\mathbf{x}}\mathbf{h}_{l}(\mathbf{x})$ is the weighted sum of the hash table entries corresponding to the nearby corners of $\mathbf{x}$.
However, the gradient $\nabla_{\mathbf{x}_k}{w_{i,l}}$ is not continuous at the corners due to the variable $s_k$, causing the direction of the gradient to flip. 
This oscillation of the gradient $\nabla_{\mathbf{x}}{w_{j,l}}$ is the source of gradient fluctuation, independently from $\mathcal{H}$.
For a detailed analysis of the derivatives and further discussion, please refer to \cref{sec:a.1.}.

%% file: 3_2_pose_refine.tex
\subsection{Camera Pose Refinement}
\label{sec:3.2.}
Camera pose can be represented as a transformation matrix from the camera coordinate to the world coordinate.
Let us denote the camera-to-world transformation matrix as $\begin{bmatrix} \mathbf{R} |   \mathbf{t} \end{bmatrix} \in \text{SE}(3)$, where $\mathbf{R} \in \text{SO}(3)$ and $\mathbf{t} \in \mathbb{R}^{3 \times 1}$ are rotation matrix and translation vector, respectively. 

\subsubsection{Pose Refinement with the Sinusoidal Encoding}
\label{sec:3.2.1.}
The pose refinement using error back-propagation in neural rendering is jointly optimizing the 6 DoF pose parameters and neural scene representation through the differentiable volume rendering:
\begin{equation}
\label{eq:joint_opt}
    \begin{split}
        \phi^{*}, \psi^{*} 
        = 
        \arg \min_{\phi, \psi} \mathcal{L}(\mathcal{I}, \hat{\mathcal{I}}; \phi, \psi),
    \end{split}
\end{equation}
where $\phi$ and $\psi$ denote model parameters and trainable camera parameters, $\hat{\mathcal{I}}$ and $\mathcal{I}$ denote reconstructed color and its ground-truth color respectively.

Note that, to our knowledge, all previous works~\cite{2020inerf,wang2021nerfmm,2022sinerf,SCNeRF2021,lin2021barf,2022garf} of pose refinement in neural rendering utilize fully differentiable encoding with respect to the input coordinate (\eg, sinusoidal or identity). 
However, they have limited performance compared to multi-resolution hash encoding~\cite{mueller2022instant}. 

\subsubsection{Pose Refinement with Multi-Resolution Hash Encoding}
\label{sec:3.2.2.}
Now, we present the optimization problem of pose refinement with multi-resolution hash encoding. 
Based on the \cref{eq:joint_opt}, we also directly optimize the camera pose parameters with multi-resolution hash encoding,
\begin{equation}
\label{eq:joint_opt2}
    \begin{split}
        \phi^{*}, \theta^{*}, \psi^{*} 
        = 
        \arg \min_{\phi, \theta, \psi} \mathcal{L}(\mathcal{I}, \hat{\mathcal{I}}; \phi, \theta, \psi),
    \end{split}
\end{equation}
where $\theta$ is a trainable parameter for multi-resolution hash encoding, \ie, the entries of the hash tables $\mathcal{H}_l$. 
However, we observe that the pose refinement and reconstruction quality from the above optimization problem is much worse than the previous works (Refer to (e) of Table~\ref{tab:ablation}). 

To explain the poor performance, we assume that the gradient fluctuation of \cref{eq:joint_opt2}, or \cref{eq:weight_grad}, negatively affects pose refinement. 
Since the input coordinate $\mathbf{x}$ is defined as a rigid transformation of the camera pose $\begin{bmatrix} \mathbf{R} |   \mathbf{t} \end{bmatrix} $ and image coordinate (projected in homogeneous space $z=-1$), the gradient fluctuation propagates through the gradient-based updates of the camera poses.
We speculate that this fluctuation makes the joint optimization of the pose refinement and the scene reconstruction difficult.
In Appendix~\ref{sec:a.2.}, we present more details of the camera pose refinement with the gradient-based optimization. 

%% file: 3_3_Smooth.tex
\subsection{Non-linear Interpolation for Smooth Gradient}
\label{sec:3.3.}
To mitigate the gradient fluctuation, 
we propose to use a \textit{smooth gradient} for the interpolation weight $w_{i,l} \in [0, 1]$ maintaining forwarding pass, inspired by the straight-through estimator~\cite{Bengio2013stochastic}.

For the {smooth gradient}, we use the activation function $\delta({w}_{i,l})$ whose derivative is zero at the corners of the hypercube, and ${w}_{i,l} \in [0, 1]$,
\begin{equation}
\label{eq:smooth_weighting}
    \begin{split}
        \delta(w_{i,l}) = 
        \frac{1- \cos(\pi w_{i,l})}{2},
    \end{split}
\end{equation}
where the activation value $\delta(w_{i,l})$ is ranged in $[0, 1]$.

As a result, the gradient of $\delta({w}_{i,l})$ with respect to $\mathbf{x}$ is derived as follows:
\begin{equation}
\label{eq:smooth_grad}
    \begin{split}
        \nabla_{\mathbf{x}}\delta({w}_{i,l})
        &= 
        \frac{\pi}{2} \sin (\pi {w}_{i,l}) 
        \cdot 
        \nabla_{\mathbf{x}}{w_{i,l}}.
    \end{split}
\end{equation}
Remind that $\nabla_{\mathbf{x}}{w_{i,l}}$ is not continuous and flipped through the boundary of a hypercube. 
In \cref{eq:smooth_grad}, the weighting by the sine function effectively makes the gradient smooth and continuous (\rf ~Figure \ref{fig:fig2}b).
Moreover, the gradient of $\mathbf{x}$ near the boundary is relatively shrunk compared to the middle of the grids, which may prevent frequent back-and-forth across the boundary after camera pose updates.

However, we do not directly use this in the interpolation forward pass. 
The cosine function in \cref{eq:smooth_weighting} unintentionally scatters the sampled points in a line toward the edges of the grids. 
This phenomenon, which we refer to as the ``zigzag problem,'' can be addressed by the \textit{straight-through} estimator~\cite{Bengio2013stochastic}. 
It maintains the results of the linear interpolation in the forward pass by the cancel-out of the last two terms in \cref{eq:straight_through}, and \emph{partially} uses the activation value $\delta({w}_{i,l})$ in the backward pass as follows:
\begin{equation}
\label{eq:straight_through}
    \begin{split}
        \hat{w}_{i,l} = w_{i,l} + \lambda \delta({w}_{i,l}) -  \lambda \tilde{\delta}({w}_{i,l}),
    \end{split}
\end{equation}
where $\lambda$ is a hyperparameter that adjusts the smooth gradient and the zigzag problem, $\tilde{\delta}$ denotes the detached variable from the computational graph. 
The steps involved in the straight-through estimator are illustrated in Figure~\ref{fig:fig1}.
\input{Figures/fig2.tex}

For an additional discussion, we present an illustration of the zigzag problem in Appendix~\ref{sec:a.3.} (See Figure~\ref{fig:zig}).
Although this straight-through estimator does not perfectly make the gradient smooth and continuous with the addition in \cref{eq:straight_through}, it is empirically more effective than other mixing variants (see Appendix~\ref{sec:a.3.} and Table~\ref{tab:lambda}).

\subsection{Curriculum Scheduling}
\label{sec:3.4.}
As argued by ~\citet{tancik2020fourier,2022pins}, NeRFs exhibit a hierarchical structure, \ie, 
the coordinate-based MLPs can suffer from spectral bias issues, in which different frequencies converge at different rates.
\citet{lin2021barf} further address this issue in the pose refinement. 
The research showed that the Jacobian of the $k^{\text{th}}$ positional encoding amplifies pose noise, making the na\"ive application of positional encoding inappropriate for pose refinement.

We observe that the multi-resolution hash encoding, which leverages the multi-level decomposition of scenes, exhibits a similar problem. 
To resolve the problem, we propose a curriculum scheduling strategy to regulate the convergence rate of the level-wise encoding.
We weight the learning rates $\eta_{l}$ of the $l^{\text{th}}$ multi-resolution hash encoding $\mathbf{h}_{l}$ by 
\begin{equation}
\label{eq:curr}
    \begin{split}
        \tilde{\eta}_{l} = r_{l}(t) \cdot \eta_{l},
    \end{split}
\end{equation}
where the weight of learning rate $r_{l}(t)$ is defined as 
\begin{equation}
\label{eq:curr_weight}
    \begin{split}
        r_{l}(t) = 
        \begin{cases}
            0 & \alpha(t) < l \\ 
            \frac{1-\cos((\alpha(t) - l) \pi)}{2} & 0 \le \alpha(t) - l < 1 \\
            1 & \text{otherwise},
        \end{cases}
    \end{split}
\end{equation}
and $\alpha(t) = L \cdot \frac{t-t_s}{t_e - t_s} \in [0, L]$ is proportional to the number of iterations $t$ in the scheduling interval $[t_s, t_e]$.

This weighting function is similar to the coarse-to-fine method proposed by \citet{park2021nerfies} and \citet{lin2021barf}. 
However, in contrast to these previous works, we apply this weighting to the learning rate of the level-wise hash table $\mathcal{H}_{l}$. 
This allows the decoding network receives the encodings from all levels, while high-level encodings are more slowly updated than the coarse levels. 
We empirically found this multi-level learning rate scheduling effective in multi-resolution hash encoding.

%% file: Figures/fig2.tex
\begin{figure}[t!]
    \begin{center}
    \centerline{\includegraphics[width=\columnwidth]{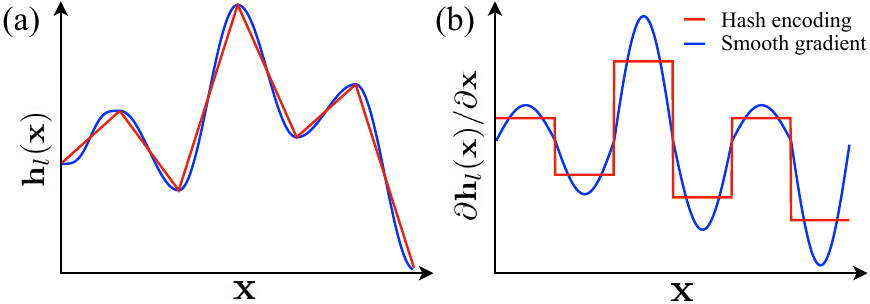}}
    \caption{Illustration on the \textit{smooth gradient} induced by \cref{eq:smooth_grad}.
    We visualize the 1D case of the multi-resolution hash encoding $\mathbf{h}_l(\mathbf{x})$ and its derivative $\partial \mathbf{h}_l(\mathbf{x}) / \partial \mathbf{x}$ in (a) and (b), respectively. 
    For further discussion, please refer to the text and Appendix~\ref{sec:a.1.}.
    }
    \label{fig:fig2}
    \end{center}
    \vskip -0.2in
\end{figure}

%% file: 4_Experiments.tex
\section{Experiment}
\label{sec:4}
In this section, we validate our proposed method using the multi-resolution hash encoding~\cite{mueller2022instant} with inaccurate or unknown camera poses. 

\input{4_1_Implementation_Details}
\input{4_2_Quantitative_Results}
\input{4_3_Qualitative_Results}

%% file: 4_1_Implementation_Details.tex
\subsection{Implementation Details}
\label{sec:4.1.}
\subsubsection{Dataset}
\label{sec:4.1.1.}
We evaluate the proposed method against the two previous works,  BARF~\cite{lin2021barf} and GARF~\cite{2022garf}.
Since the implementation of GARF is unavailable, we re-implement GARF. 
Our re-implemented GARF has the same structure as BARF except for sinusoidal encoding and Gaussian activation.
Following \citet{lin2021barf} and \citet{2022garf}, we evaluate and compare our method on two public novel-view-synthesis datasets. 

\paragraph{NeRF-Synthetic.}
{NeRF-Synthetic}~\cite{mildenhall2020nerf} has 8 synthetic object-centric scenes, which consist of $100$ rendered images with ground-truth camera poses (intrinsic and extrinsic) for each scene. 
Following \citet{lin2021barf}, we utilize this dataset for the noisy camera pose scenario.
To simulate the scenario of imperfect camera poses, we adopt the approach in \citet{lin2021barf} synthetically perturbing the camera poses with additive Gaussian noise, $\delta \psi \sim \mathcal{N}(\mathbf{0}, \ 0.15 \mathbf{I})$.

\paragraph{LLFF.}
{LLFF}~\cite{mildenhall2019llff} has 8 forward-facing scenes captured by a hand-held camera, including RGB images and camera poses that have been estimated using the off-the-shelf algorithm~\cite{2016sfm}. 
Following previous works, we utilize this dataset for the \textit{unknown} camera pose scenario. 
Unlike the synthetic datasets, we initialize all camera poses with the \textit{identity} matrix. 
Note that, the camera poses provided by \textbf{LLFF} are the estimations obtained using the COLMAP algorithm~\cite{2016sfm}. 
As such, the pose error measured in our quantitative results only indicates the agreement between the learned pose and the estimated pose using the classical geometry-based approach.

\subsubsection{Implementation Details}
\label{sec:4.1.2.}
For the multi-resolution hash encoding, we follow the approach of Instant-NGP~\cite{mueller2022instant}, which uses a table size of $T = 2^{19}$ and a dimensionality of $F=2$ for each level feature. 
Each feature table is initialized with a uniform distribution $\mathcal{U}[0,\ 1e-4]$.
Note that we reproduce the entire training pipeline in PyTorch for pose refinement instead of using the original C$++$ \& CUDA implementation of Instant-NGP for fair comparison~\footnote{While our re-implementation performs almost the same with the original, it takes slightly longer training time due to PyTorch's execution latency. The performance of our re-implemented Instant-NGP is reported in Appendix~\ref{sec:b.1.}.}.

The decoding network consists of 4-layer MLPs with ReLU~\cite{2011relu} activation and 256 hidden dimensions, including density network branch and color. 
We utilize the tiny-cuda-nn (tcnn)~\cite{2022tcnn} framework for the decoding network. 
We present the other implementation details in Appendix~\ref{sec:b.1.}.
While we set $\lambda = 1$ by default for the \textit{straight-through estimator}, the other options are explored in Appendix~\ref{sec:a.3.}.

\subsubsection{Evaluation Criteria}
\label{sec:4.1.3.}
In conformity with previous studies~\cite{lin2021barf,2022garf}, we evaluate the performance of our experiments in two ways: 1) the quality of view-synthesis for the 3D scene representation and 2) the accuracy of camera pose registration.
We measure the PSNR, SSIM, and LPIPS scores for view-synthesis quality, as employed in the original NeRF~\cite{mildenhall2020nerf}. 
The rotation and translation errors are defined as follows: 
\begin{align}
\label{eq:pose_error}
    &E(\mathbf{R}) = \cos^{-1}\Big(\frac{\textit{tr}(\mathbf{R}'\cdot \mathbf{R}^{\rm T})-1}{2}\Big),
    \\
    &E(\mathbf{t}) \ = |\mathbf{t}' - \mathbf{t}|^2_2,
\end{align}
where $\begin{bmatrix} \mathbf{R}' |   \mathbf{t}' \end{bmatrix} \in \text{SE}(3)$ denotes the ground-truth camera-to-world transformation matrix and $\textit{tr}(\cdot)$ denotes trace operator.
Like the \citet{lin2021barf}, all the metrics are measured after the pre-alignment stage using the Procrustes analysis.
In experiments, all the camera poses $\psi$ are parameterized by the \se ~Lie algebra with known intrinsics. 

%% file: 4_2_Quantitative_Results.tex
\subsection{Quantitative Results}
\input{Tables/synthetic}

\input{Tables/llff}

\subsubsection{Synthetic Objects in NeRF-Synthetic}
\label{sec:4.2.1}
Table~\ref{tab:synthetic} demonstrates the quantitative results of the NeRF-Synthetic. 
In Table~\ref{tab:synthetic}, the proposed method achieves state-of-the-art performances in both pose registration and reconstruction fidelity across all scenes.
The results align with \citet{mueller2022instant} showing impressive performance on the scenes with high geometric details.

On the other hand, \citet{mueller2022instant} previously demonstrated that multi-resolution hash encoding is limited to the scenes with complex and view-dependent reflections, \ie, \textit{Materials}. 
Although they attributed this limitation to their shallow decoding networks, we observed similar performance when utilizing deeper decoding networks.
We hypothesize that frequency-based encodings, such as sinusoidal or spherical harmonics, might be more appropriate for addressing complex and view-dependent reflections. 
We will further investigate this issue in future work.

\subsubsection{Real-World Scenes in LLFF}
\label{sec:4.2.2}
We report the quantitative results of the LLFF dataset in Table~\ref{tab:llff}. 
Note that GARF utilizes 6-layer decoding networks for this dataset.
In Table~\ref{tab:llff},  the proposed method outperforms the previous methods regarding reconstruction fidelity and pose recovery, especially for translation.
These results suggest that the learned pose from our method is closely related to that of the classical geometric algorithm, indicating that our proposed method can learn camera poses from scratch using the multi-resolution hash encoding.

In terms of rotation angle registration, our method outperforms BARF, achieving comparable performance to GARF. 
Still, notice that our method achieves the best view-synthesis quality compared to the other methods.
Also, in Table~\ref{tab:llff_colmap}, we investigate the interaction with the COLMAP camera pose initialization and our method. 
Please refer to \cref{sec:b.2.} for the details.
Here, the underbar denotes runners-up. 

\subsection{Ablation Study}
\label{sec:4.2.3}
We present additional ablation studies to examine the proposed method's effectiveness. 
Similar to the Instant-NGP~\cite{mueller2022instant}, all the following experiments are conducted on the \textit{Hotdog} in the NeRF-Synthetic dataset for comparison. 
Note that other scenes behave similarly. 

\input{Tables/ablations.tex}
\input{Tables/llff_colmap.tex}
\subsubsection{Component Analysis}
In Table~\ref{tab:ablation}, we perform the ablation study for our method to examine the role of each element.
As shown in row (b) compared with (c), the \textit{smooth gradient} significantly helps with pose refinements, resulting in more accurate pose registration and higher view-synthesis quality. 
Also, from (a) and (b), we observe that the \textit{straight-through estimator} prevents unintentional jittering from the non-linear weighting showing outperformance.
Lastly, as shown in (a) and (d), our proposed multi-level learning rate scheduling reasonably enhances pose estimation and scene reconstruction qualities.

\subsubsection{Time Complexity}
In Figure~\ref{fig:fig3}, we visualize the comparison of the training speed between the proposed method and the previous works~\cite{lin2021barf,2022garf}.
By utilizing fast convergence of multi-resolution hash encoding, the proposed method achieves more than $20\times$ faster training speed compared to the previous works. 
Remind that the proposed method outperforms previous methods both in pose registration and view synthesis. 
\input{Figures/fig3.tex}

\subsubsection{Decoder Size}
Here, we examine the design criteria for decoding networks $m(\mathbf y; \phi)$ in terms of model capacity. 
The original implementation of Instant-NGP~\cite{mueller2022instant} utilizes shallow decoding networks, resulting in the feature table $\mathcal{H}$ having a relatively larger number of learnable parameters than the decoding networks, \ie, $|\theta| \gg |\phi|$.
We find that this often leads to the suboptimal convergence of both the multi-resolution hash encoding and the camera pose registration. 

Figure~\ref{fig:fig4} presents the view-synthesis quality with respect to varying model sizes of the decoding network.
Unlike the findings of \citet{mueller2022instant}, who did not observe improvement with deeper decoder MLPs (as shown by the dashed line in the Figure), we observe that the decoder size heavily impacts both the view synthesis and the pose registration. 
Therefore, in cases where the camera pose is inaccurate, we assume a sufficient number of parameters in the decoder is necessary.
Informed by this analysis, we employ deeper and wider decoding networks than the original Instant-NGP: 4-layer MLPs with 256 neurons. 
Note that competitive methods~\cite{lin2021barf,2022garf} utilize a deeper decoder network with 8-layer MLPs with 256 neurons having more parameters.
\input{Figures/fig4.tex}

%% file: Tables/synthetic.tex
\begin{table*}[t]
  \centering
  \small
  \vspace{-0.7em}
  \caption{
  \text{Quantitative results of the NeRF-Synthetic dataset.}
  }
  \label{tab:synthetic}
  \vskip 0.13in
  \begin{adjustbox}{width=1\textwidth} 
  \begin{tabular}{ c  c c c c c c   c c c c  c c c c c c c c }
    \toprule
    
    \multirow{4}{*}{Scene} 
    & \multicolumn{6}{c}{Camera Pose Registration} 
    & \multicolumn{9}{c}{View Synthesis Quality} \\ 
    
    \cmidrule(r){2-7}  
    \cmidrule(r){8-16} 
    & \multicolumn{3}{c}{Rotation $({}^{\circ})$ $\downarrow$}
    & \multicolumn{3}{c}{Translation $\downarrow$}
    & \multicolumn{3}{c}{PSNR $\uparrow$}
    & \multicolumn{3}{c}{SSIM $\uparrow$}
    & \multicolumn{3}{c}{LPIPS $\downarrow$} \\
    \cmidrule(r){2-4}
    \cmidrule(r){5-7}
    \cmidrule(r){8-10}
    \cmidrule(r){11-13}
    \cmidrule(r){14-16}  
    & GARF
    & BARF
    & Ours
    & GARF
    & BARF
    & Ours
    & GARF
    & BARF
    & Ours
    & GARF
    & BARF
    & Ours
    & GARF
    & BARF
    & Ours
    \\
    \midrule 
    
    Chair 
    & 0.113
    & 0.096
    & \textbf{0.085}
    
    & 0.549
    & 0.428
    & \textbf{0.365}
    
    & 31.32
    & 31.16
    & \textbf{31.95}
    
    & 0.959
    & 0.954
    & \textbf{0.962}
    
    & 0.042
    & 0.044
    & \textbf{0.036}
    \\
    
    Drum 
    & 0.052
    & 0.043
    & \textbf{0.041}
    
    & 0.232
    & 0.225
    & \textbf{0.214}
    
    & 24.15
    & 23.91
    & \textbf{24.16}
    
    & 0.909
    & 0.900
    & \textbf{0.912}
    
    & 0.097
    & 0.099
    & \textbf{0.087}
    \\
    
    Ficus 
    & 0.081
    & 0.085
    & \textbf{0.079}
    
    & \textbf{0.461}
    & {0.474}
    & 0.479
    
    & 26.29
    & 26.26
    & \textbf{28.31}
    
    & 0.935
    & 0.934
    & \textbf{0.943}
    
    & 0.057
    & 0.058
    & \textbf{0.051}
    \\
    
    Hotdog 
    & 0.235
    & 0.248
    & \textbf{0.229}
    
    & \textbf{1.123}
    & 1.308
    & \textbf{1.123}
    
    & 34.69
    & 34.54
    & \textbf{35.41}
    
    & 0.972
    & 0.970
    & \textbf{0.981}
    
    & 0.029
    & 0.032
    & \textbf{0.027}
    \\
    
    Lego 
    & 0.101 
    & 0.082
    & \textbf{0.071}
    
    & 0.299
    & 0.291
    & \textbf{0.272}
    
    & 29.29
    & 28.33
    & \textbf{31.65}
    
    & 0.925
    & 0.927
    & \textbf{0.973}
    
    & 0.051
    & 0.050
    & \textbf{0.036}
    \\
    
    Materials 
    & \textbf{0.842}
    & {0.844}
    & 0.852
    
    & \textbf{2.688}
    & {2.692}
    & 2.743
    
    & \textbf{27.91}
    & {27.84}
    & 27.14
    
    & \textbf{0.941}
    & 0.936 
    & 0.911
    
    & 0.059
    &\textbf{0.058}
    & 0.062
    \\
    
    Mic 
    & 0.070
    & 0.071
    & \textbf{0.068}
    
    & 0.293
    & 0.301
    & \textbf{0.287}
    
    & 31.39
    & 31.18
    & \textbf{32.33}
    
    & 0.971
    & 0.969
    & \textbf{0.975}
    
    & 0.047
    & 0.048
    & \textbf{0.043}
    \\
    
    Ship 
    & \textbf{0.073}
    & {0.075}
    & 0.079
    
    & 0.310
    & 0.326
    & \textbf{0.287}
    
    & 27.64
    & 27.50
    & \textbf{27.92}
    
    & 0.862
    & 0.849
    & \textbf{0.879}
    
    & 0.119
    & 0.132
    & \textbf{0.110}
    \\
    
    \midrule
    Mean 
    & 0.195
    & 0.193
    & \textbf{0.189}
    
    & 0.744
    & 0.756
    & \textbf{0.722}
    
    & 28.96
    & 28.84
    & \textbf{29.86}
    
    & 0.935
    & 0.930
    & \textbf{0.943}
    
    & 0.063
    & 0.065
    & \textbf{0.056}
    \\
    
    \bottomrule
  \end{tabular}
  \end{adjustbox}
  \vskip -0.1in
\end{table*}

%% file: Tables/llff.tex
\begin{table*}[t]
  \centering
  \small
  \caption{
      Quantitative results of the LLFF dataset.
  }
  \label{tab:llff}
  \vskip 0.13in
  \begin{adjustbox}{width=1\textwidth} 
  \begin{tabular}{ c  c c c c c c   c c c c  c c c c c c c c }
    \toprule
    
    \multirow{4}{*}{Scene} 
    & \multicolumn{6}{c}{Camera Pose Registration} 
    & \multicolumn{9}{c}{View Synthesis Quality} \\ 
    
    \cmidrule(r){2-7}  
    \cmidrule(r){8-16} 
    & \multicolumn{3}{c}{Rotation $({}^{\circ})$ $\downarrow$}
    & \multicolumn{3}{c}{Translation $\downarrow$}
    & \multicolumn{3}{c}{PSNR $\uparrow$}
    & \multicolumn{3}{c}{SSIM $\uparrow$}
    & \multicolumn{3}{c}{LPIPS $\downarrow$} \\
    \cmidrule(r){2-4}
    \cmidrule(r){5-7}
    \cmidrule(r){8-10}
    \cmidrule(r){11-13}
    \cmidrule(r){14-16}  
    & GARF
    & BARF
    & Ours
    & GARF
    & BARF
    & Ours
    & GARF
    & BARF
    & Ours
    & GARF
    & BARF
    & Ours
    & GARF
    & BARF
    & Ours
    \\
    \midrule 
    
    Fern 
    & 0.470
    & 0.191
    & \textbf{0.110}
    
    & 0.250
    & \textbf{0.102}
    & \textbf{0.102}
    
    & 24.51
    & 23.79
    & \textbf{24.62}
    
    & 0.740
    & 0.710
    & \textbf{0.743}
    
    & 0.290
    & 0.311
    & \textbf{0.285}
    \\
    
    Flower 
    & 0.460
    & \textbf{0.251} 
    & 0.301
    
    & 0.220
    & 0.224
    & \textbf{0.211}
    
    & \textbf{26.40}
    & 23.37
    & 25.19
    
    & \textbf{0.790}
    & 0.698
    & 0.744
    
    & \textbf{0.110}
    & 0.211
    & 0.128
    \\
    
    Fortress 
    & \textbf{0.030}
    & 0.479 
    & 0.211
    
    & 0.270
    & 0.364
    & \textbf{0.241}
    
    & 29.09
    & 29.08
    & \textbf{30.14}
    
    & 0.820
    & 0.823
    & \textbf{0.901}
    
    & 0.150
    & 0.132
    & \textbf{0.098}
    \\
    
    Horns 
    & \textbf{0.030}
    & 0.304 
    & 0.049
    
    & 0.210 
    & 0.222 
    & \textbf{0.209}
    
    & 22.54
    & 22.78
    & \textbf{22.97}
    
    & 0.690
    & 0.727
    & \textbf{0.736}
    
    & 0.330
    & 0.298
    & \textbf{0.290}
    \\
    
    Leaves 
    & \textbf{0.130}
    & 1.272
    & 0.840
    
    & 0.230
    & 0.249
    & \textbf{0.228}
    
    & \textbf{19.72}
    & 18.78
    & 19.45
    
    & \textbf{0.610}
    & 0.537
    & 0.607
    
    & 0.270
    & 0.353
    & \textbf{0.269}
    \\
    
    Orchids 
    & 0.430
    & 0.627
    & \textbf{0.399}
    
    & 0.410
    & 0.404
    & \textbf{0.386}
    
    & 19.37
    & 19.45
    & \textbf{20.02}
    
    & 0.570
    & 0.574
    & \textbf{0.610}
    
    & 0.260
    & 0.291
    & \textbf{0.213}
    \\
    
    Room 
    & \textbf{0.270}
    & 0.320
    & 0.271
    
    & \textbf{0.200}
    & 0.270
    & 0.213
    
    & 31.90
    & 31.95
    & \textbf{32.73}
    
    & 0.940
    & {0.949}
    & \textbf{0.968}
    
    & 0.130
    & {0.099}
    & \textbf{0.098}
    \\
    
    T-Rex 
    & \textbf{0.420}
    & 1.138 
    & 0.894
    
    & \textbf{0.360}
    & 0.720 
    & 0.474
    
    & 22.86
    & 22.55
    & \textbf{23.19}
    
    & 0.800
    & 0.767
    & \textbf{0.866}
    
    & 0.190
    & 0.206
    & \textbf{0.183}
    \\
    
    \midrule
    Mean 
    & \textbf{0.280}
    & 0.573
    & 0.384
    
    & 0.269
    & 0.331
    & \textbf{0.258}
    
    & 24.55
    & 23.97
    & \textbf{24.79}
    
    & 0.745
    & 0.723
    & \textbf{0.772}
    
    & 0.216
    & 0.227
    & \textbf{0.197}
    \\
    
    \bottomrule
  \end{tabular}
  \end{adjustbox}
  \vskip -0.1in
\end{table*}

%% file: Tables/ablations.tex
     
    
    
     

    

\begin{table*}[!ht]
  \centering
  \small
  \vspace{-0.7em}
  \caption{Ablation study on the components of the proposed method. Experiments are conducted on the \textit{Hotdog} in the NeRF-Synthetic dataset.
  Three components are the straight-through estimator in \cref{eq:straight_through}, the smooth gradient with cosine activation in \cref{eq:smooth_weighting}, and the curriculum scheduling in \cref{sec:3.4.}. 
  }
  \label{tab:ablation}
  \vskip 0.15in
  \begin{tabular}{ c  c c c  c c c  }
    \toprule
    \multirow{2}{*}{} 
    & \multicolumn{3}{c}{\textbf{Component Ablation}} 
    & \multicolumn{3}{c}{\textbf{Evaluation Metric}} 
    \\
    \cmidrule{2-4} 
    \cmidrule{5-7}
    & $\ $\textit{w/ Straight-Through} $\ $
    & $\ $\textit{w/ Smooth Grad.} $\ $
    & $\ $\textit{w/ Curriculum Scheduling} $\ $\hspace{1pt} 
    & {Rotation $({}^{\circ})$ $\downarrow$} 
    & {Translation $\downarrow$} 
    & {PSNR $\uparrow$} 
    \\  
    \midrule
     
    (a)
    & \checkmark & \checkmark & \checkmark 
    & \textbf{0.234}
    & \textbf{1.124}
    & \textbf{35.41}
    \\ 
    
    (b)
    &  & \checkmark & \checkmark
    & 0.245
    & 1.130
    & 35.03
    \\
    
    (c)
    &  &  & \checkmark
    & 0.977
    & 3.210
    & 29.89
    \\
     
    (d)
    & \checkmark & \checkmark & 
    & 0.447
    & 1.921
    & 32.19
    \\

    (e)
    &  &  & 
    & 2.779
    & 6.423
    & 25.41
    \\
    
    \bottomrule
  \end{tabular}
  \vspace{1pt}
\end{table*}

%% file: Tables/llff_colmap.tex
\begin{table*}[t!]
  \centering
  \small
  \vspace{-0.7em}
  \caption{
      Quantitative results of the proposed method in the LLFF dataset with the COLMAP initialization (PSNR $\uparrow$).
  }
  \label{tab:llff_colmap}
  \vskip 0.15in
  \begin{adjustbox}{width=1\textwidth} 
  \begin{tabular}{c  c c  c c c c c c c c  c }
    \toprule
    & \multicolumn{2}{c}{Experimental Setting}
    & \multicolumn{9}{c}{LLFF}  \\ 
    \cmidrule(r){2-3}   
    \cmidrule(r){4-12}   
    
    & \textit{w/} COLMAP 
    & \textit{w/} Pose Refinement
    & Fern
    & Flower
    & Fortress
    & Horns
    & Leaves
    & Orchids
    & Room
    & T-Res
    & \textit{Average}
    \\
    \midrule 
    (a)
    & \checkmark
    & 
    & \underbar{25.83}
    & \underbar{26.56}
    & 28.00
    & \underbar{26.46}
    & 18.89
    & \underbar{20.15}
    & 31.96
    & \underbar{26.51}
    & \underbar{25.55}
    \\
    (b)
    &
    & \checkmark
    & 24.62 
    & 25.19
    & \underbar{30.14}
    & 22.97
    & \underbar{19.45}
    & 20.02
    & \underbar{32.73}
    & 23.19
    & 24.79
    \\
    (c)
    & \checkmark
    & \checkmark
    & \textbf{26.41}
    & \textbf{28.00}
    & \textbf{30.99}
    & \textbf{27.35}
    & \textbf{19.97}
    & \textbf{21.26}
    & \textbf{33.02}
    & \textbf{26.83}
    & \textbf{26.73}
    \\
    
    \bottomrule
  \end{tabular}
  \end{adjustbox}
  \vskip -0.1in
\end{table*}


%% file: Figures/fig3.tex
\begin{figure}[t]
    \begin{center}
    \centerline{\includegraphics[width=\columnwidth]{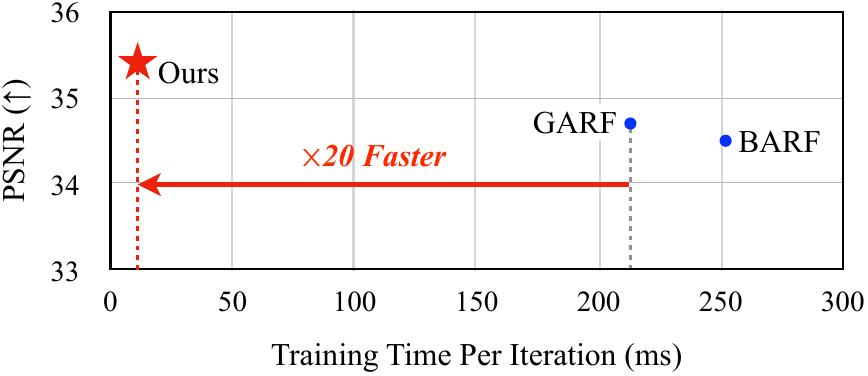}}
    \vskip -0.5em
    \caption{Comparison of the averaged training time per iteration on the \textit{Hotdog} in the NeRF-Synthetic dataset.
    Our method takes only 10.8 ms,  significantly faster than the previous works, GARF and BARF, which are 213 ms and 252 ms, respectively.
    }
    \label{fig:fig3}
    \end{center}
    \vskip -0.2in
    \vskip -0.5em
\end{figure}

%% file: Figures/fig4.tex
\begin{figure}[t!]
    \begin{center}
    \centerline{\includegraphics[width=0.95\columnwidth]{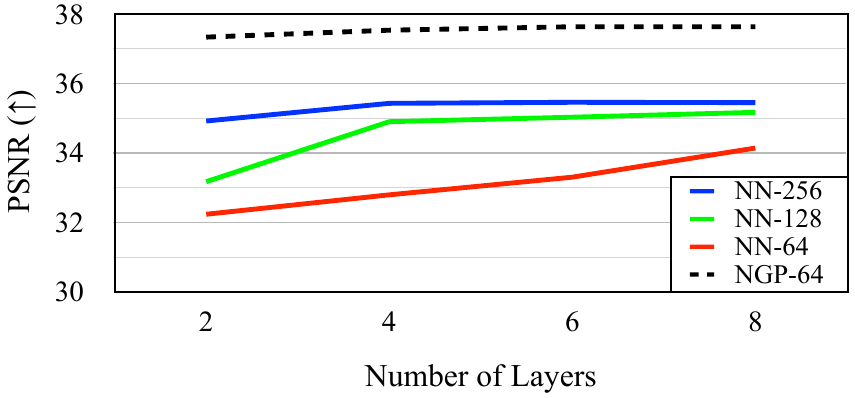}}
    \vskip -0.5em
    \vskip -0.15em
    \caption{Performance depends on the decoder size. 
    We plot as the depth of the decoder increases, varying the hidden size from 64 to 256.
    The dashed line denotes the NGP's with the hidden size of 64 using the ground-truth camera poses as the upper bound.
    }
    \label{fig:fig4}
    \end{center}
    \vskip -1.3em
\end{figure}

%% file: 4_3_Qualitative_Results.tex
\subsection{Qualitative Results}
\label{subsubsec:4.3.}
We present the qualitative results of our method compared with competitive methods. Please refer to Appendix~\ref{sec:b.4.}.

%% file: 5_Conclusion.tex
\section{Conclusion}
We investigate the joint optimization of camera poses and scene reconstruction using multi-resolution hash encoding.
Based on the careful analysis of the gradient fluctuation of the hash encoding, we propose a simple yet effective method of the straight-through estimator for gradient smoothing.
Additionally, we consider the spectral bias of multi-level decomposition and adopt a multi-level learning rate scheduling varying convergence rates of the multi-level encodings.
Our extensive experiments show that the proposed method successfully recovers the camera poses with state-of-the-art performance on NeRF-Synthetic and LLFF.
However, it shows limited performance in the scenes with complex reflections, inherited from the multi-resolution hash encoding~\cite{mueller2022instant}.
Nevertheless, with its state-of-the-art performance and fast convergence, we believe our method is a reasonable choice for real-world problems with imperfect or unknown camera poses. 

%% file: Appendix.tex
\newpage
\appendix
\onecolumn
\input{Appendix_A.tex}
\input{Appendix_B.tex}

%% file: Appendix_A.tex
\section{The Derivative of Multi-Resolution Hash Encoding}
\label{sec:a}
We describe the details of the gradient-based optimization of the camera poses in the multi-resolution hash encoding.

\subsection{Jacobian of the Interpolated Feature Vector}
\label{sec:a.1.}
Borrowed from the \cref{eq:grad} in the manuscript, the Jacobian $\nabla_{\mathbf{x}}\mathbf{h}_{l}(\mathbf{x}) \in \mathbb{R}^{F \times d}$ of the $l^{\text{th}}$ interpolated feature vector $\mathbf{h}_l (\mathbf{x})$ with respect to $\mathbf{x}$ can be derived using the chain-rule as follows: 
\begin{equation}
\label{eq:grad_re}
    \begin{split}
        \nabla_{\mathbf{x}}\mathbf{h}_{l}(\mathbf{x}) 
        &= 
        \left[  
            \frac{\partial {\mathbf{h}_{l}}(\mathbf{x})}{\partial {x}_1}, 
            \dots,
            \frac{\partial {\mathbf{h}_{l}}(\mathbf{x})}{\partial {x}_d}
        \right]
        \\
        &= 
        \sum_{i=1}^{2^{d}} 
        \mathcal{H}_{l}\big( h_{l} \big( \mathbf{c}_{i,l} (\mathbf{x}) \big) \big)
        \cdot
        \left[  
            \frac{\partial {{w}_{i,l}}(\mathbf{x})}{\partial {x}_1}, 
            \dots,
            \frac{\partial {{w}_{i,l}}(\mathbf{x})}{\partial {x}_d}
        \right].
    \end{split}
\end{equation}

Let $\bar{i}$ be one of the nearest corner indices from $\mathbf{c}_{i,l}$ in a unit hypercube, where $\mathbf{c}_{i,l}$ and $\mathbf{c}_{\bar{i},l}$ make an edge of the unit hypercube.
Among the $2^d$ corners, we have $2^{d-1}$ pairs like that. 
Then, we have the relation for $w_{\bar{i}_k,l}$ as follows:
\begin{equation}
\label{eq:conjugate}
    \begin{split}
        \frac{\partial {{w}_{\bar{i}_k,l}}(\mathbf{x})}{\partial {x}_k} 
        = 
        - \frac{\partial {{w}_{i,l}}(\mathbf{x})}{\partial {x}_k},
    \end{split}
\end{equation}
which can be inferred from \cref{eq:weight_grad} since the relative positions of $\mathbf{x}$ are different for the two cases.

Using this relation and the appropriate choice of the indices, the $k^{\text{th}}$ element of Jacobian $\nabla_{\mathbf{x}}\mathbf{h}_{l}(\mathbf{x})$ can be rewritten as follows:
\begin{equation}
\label{eq:grad2}
    \begin{split}
        \frac{\partial {\mathbf{h}_{l}}(\mathbf{x})}{\partial {x}_k}
        &= 
        \sum_{i=1}^{2^{d}} 
        \mathcal{H}_{l}\big( h_{l} \big( \mathbf{c}_{i,l} (\mathbf{x}) \big) \big)
        \cdot
        \frac{\partial {{w}_{i,l}}(\mathbf{x})}{\partial {x}_k}
        \\
        &=
        \sum_{i=1}^{2^{d-1}} 
        \left(
        \mathcal{H}_{l}\big( h_{l} \big( \mathbf{c}_{i,l} (\mathbf{x}) \big) \big)
        -
        \mathcal{H}_{l}\big( h_{l} \big( \mathbf{c}_{\bar{i}_k,l} (\mathbf{x}) \big) \big)
        \right)
        \cdot
        \frac{\partial {{w}_{i,l}}(\mathbf{x})}{\partial {x}_k} 
        \\ 
        &=
        \sum_{i=1}^{2^{d-1}} 
        \left(
        \mathcal{H}_{l}\big( h_{l} \big( \mathbf{c}_{i,l} (\mathbf{x}) \big) \big)
        -
        \mathcal{H}_{l}\big( h_{l} \big( \mathbf{c}_{\bar{i}_k,l} (\mathbf{x}) \big) \big)
        \right)
        \cdot
        \prod_{j \neq k} 
        \left(
            1 - | \mathbf{x}_{l} - \mathbf{c}_{i,l}(\mathbf{x}) |_j 
        \right ),
    \end{split}
\end{equation}
where $\prod_{j \neq k} \left( 1 - | \mathbf{x}_{l} - \mathbf{c}_{i,l}(\mathbf{x}) |_j \right )$ and the differences between the hash table entries are constant to the $x_k$, which make ${\partial {\mathbf{h}_{l}}(\mathbf{x})} / {\partial {x}_k}$ is constant along with the $k^{\text{th}}$ axis of the unit hypercube. 
Notice that the last term can be seen as the weights defined as:
\begin{equation}
    \sum_{i=1}^{2^{d-1}} \prod_{j \neq k} \left( 1 - | \mathbf{x}_{l} - \mathbf{c}_{i,l}(\mathbf{x}) |_j \right ) = 1,
\end{equation}
where ${\partial {\mathbf{h}_{l}}(\mathbf{x})} / {\partial {x}_k}$ is the convex combination of the differences between two hash table entries.

Our speculation from this analysis is that ${\partial {\mathbf{h}_{l}}(\mathbf{x})} / {\partial {x}_k}$ is the linear slope defined by the differences among the hash table entries consist of each grid.
However, the slope is sharply changed when crossing the boundary of the grid. 
For the one-dimensional case, this is a kind of \textit{triangular wave}~\footnote{\url{https://en.wikipedia.org/wiki/Triangle_wave}}, which is a periodic, piecewise linear, continuous real function. 
This function has sharp local minima and maxima at the vertices of the wave, hindering appropriate gradient steps for accurate pose refinement.

Our method is proposed to remedy this problem. 
We empirically show that the smaller gradient for the positions near the boundaries compared with the other positions is significantly helpful for accurate pose refinement.

\subsection{Camera-to-World Transformation}
\label{sec:a.2.}
This section shows that the gradient fluctuation of the Jacobian $\nabla_{\mathbf{x}}\mathbf{h}_{l}(\mathbf{x})$ also affects the gradient-based optimization of the camera poses by the chain-rule. 

Remind that we denote the camera-to-world transformation matrix as $\begin{bmatrix} \mathbf{R} |   \mathbf{t} \end{bmatrix} \in \text{SE}(3)$, where $\mathbf{R} \in \text{SO}(3)$ and $\mathbf{t} \in \mathbb{R}^{3 \times 1}$, which are rotation matrix and translation vector, respectively. 
Typically, the translation vector $\mathbf{t}$ is expressed as a 3D vector in Euclidean space. 
Whereas for the rotation matrix on $\text{SO}(3)$ space, previous works adopt the 3 DoF axis-angle representation with the Rodrigues’ formula or the 6D continuous representation~\cite{zhou2018rot,SCNeRF2021}.

Consider an input coordinate of the image space as $\mathbf{x}_{I} \in \mathbb{R}^2$. 
To transform the image coordinate to the world coordinate, the image coordinate is projected to its homogeneous space $z = -1$. 
Then, we get the corresponding world coordinate $\mathbf{x} \in \mathbb{R}^{3 \times 1}$ by the rigid transformation of the camera pose as follows: 
\begin{equation}
\label{eq:pose_transform}
    \begin{split}
        \mathbf{x} = \mathbf{R} \ { \begin{bmatrix} \mathbf{x}_{I} \\ -1  \end{bmatrix}} + \mathbf{t}.
    \end{split}
\end{equation}
In volume rendering, the emission-absorption ray casting utilizes the transformed coordinate $\mathbf{x}$ to sample the points $\mathbf{x'}$ in a casted ray.
Therefore, the derivative of camera pose parameters with respect to the $\mathbf{x'}$ is derived as follows: 
\begin{align}
\label{eq:cam_grad}
    & \nabla_{\mathbf{R}}\mathbf{h}_{l}(\mathbf{x'})
    = 
    \frac{\partial \mathbf{h}_l (\mathbf{x})}{\partial \mathbf{x'}}
    \cdot
    \frac{\partial \mathbf{x'}}{\partial \mathbf{R}},  
    \\ 
    & \nabla_{\mathbf{t}}\mathbf{h}_{l}(\mathbf{x'})
    = 
    \frac{\partial \mathbf{h}_l (\mathbf{x'})}{\partial \mathbf{x'}}
    \cdot
    \frac{\partial \mathbf{x'}}{\partial \mathbf{t}}.
\end{align}
As shown in the above equation, the relations between $\nabla_{\mathbf{x'}}\mathbf{h}_{l}(\mathbf{x'})$ and, the Jacobian of $\nabla_{\mathbf{R}}\mathbf{h}_{l}(\mathbf{x'})$ and $\nabla_{\mathbf{R}}\mathbf{h}_{l}(\mathbf{x})$ are linear. 
Therefore, the gradient fluctuation of the $\nabla_{\mathbf{x'}}\mathbf{h}_{l}(\mathbf{x'})$ directly propagates to the gradient with respect to the camera poses, resulting in inappropriate updates for the pose refinement.

\paragraph{Comparison with Sinusoidal Encoding.}
The main difference of the pose refinement using multi-resolution hash encoding from sinusoidal encoding is \textit{unsmoothness}.
In the case of the sinusoidal encoding~\cite{lin2021barf,SCNeRF2021}, the encoding output of the \cref{eq:trilinear} is replaced with a differentiable encoding function $\gamma(\mathbf{x})$.
Rather than linearly approximating the feature of $\mathbf{x}$ within a unit hypercube, it is directly encoded using the sinusoidal encoding as follows: 
\begin{equation}
\label{eq:sinusoidal}
    \begin{split}
        \mathbf{h}_{l}(\mathbf{x}) 
        = \gamma_{l}(\mathbf{x}) 
        = 
        \left[
            \cos(2^l \pi \mathbf{x}),\ \sin(2^l \pi \mathbf{x}) 
        \right].
    \end{split}
\end{equation}
Unlike the multi-resolution hash encoding, it induces the smooth gradient with respect to the input coordinate $\mathbf{x}$. 

\subsection{Zigzag Problem and the Straight-Through Estimator}
\label{sec:a.3.}
As discussed in the manuscript, if we use smooth interpolation weight directly, the cosine function in \cref{eq:smooth_weighting} unintentionally scatters the sampled points, which we refer to as the zigzag problem. 
We present an illustration of the zigzag problem in Figure~\ref{fig:zig}. 
As shown on the right side of the figure, if we directly use the smooth interpolation in a hypercube, the sampled points get closer to the edges. \newline \newline
To remedy this problem, we leverage the straight-through estimator where the hyperparameter $\lambda$ adjusts the ``zigzag problem'' between the original weight and the activation value.
We explore the $\lambda$ in Table~\ref{tab:lambda} to investigate this impact.

\input{Figures/zigzag.tex}
\input{Tables/lambda.tex}

%% file: Figures/zigzag.tex
\begin{figure}[hbt!]
    \begin{center}
    \begin{minipage}{0.9\textwidth}
    \centerline{\includegraphics[width=0.618\columnwidth]{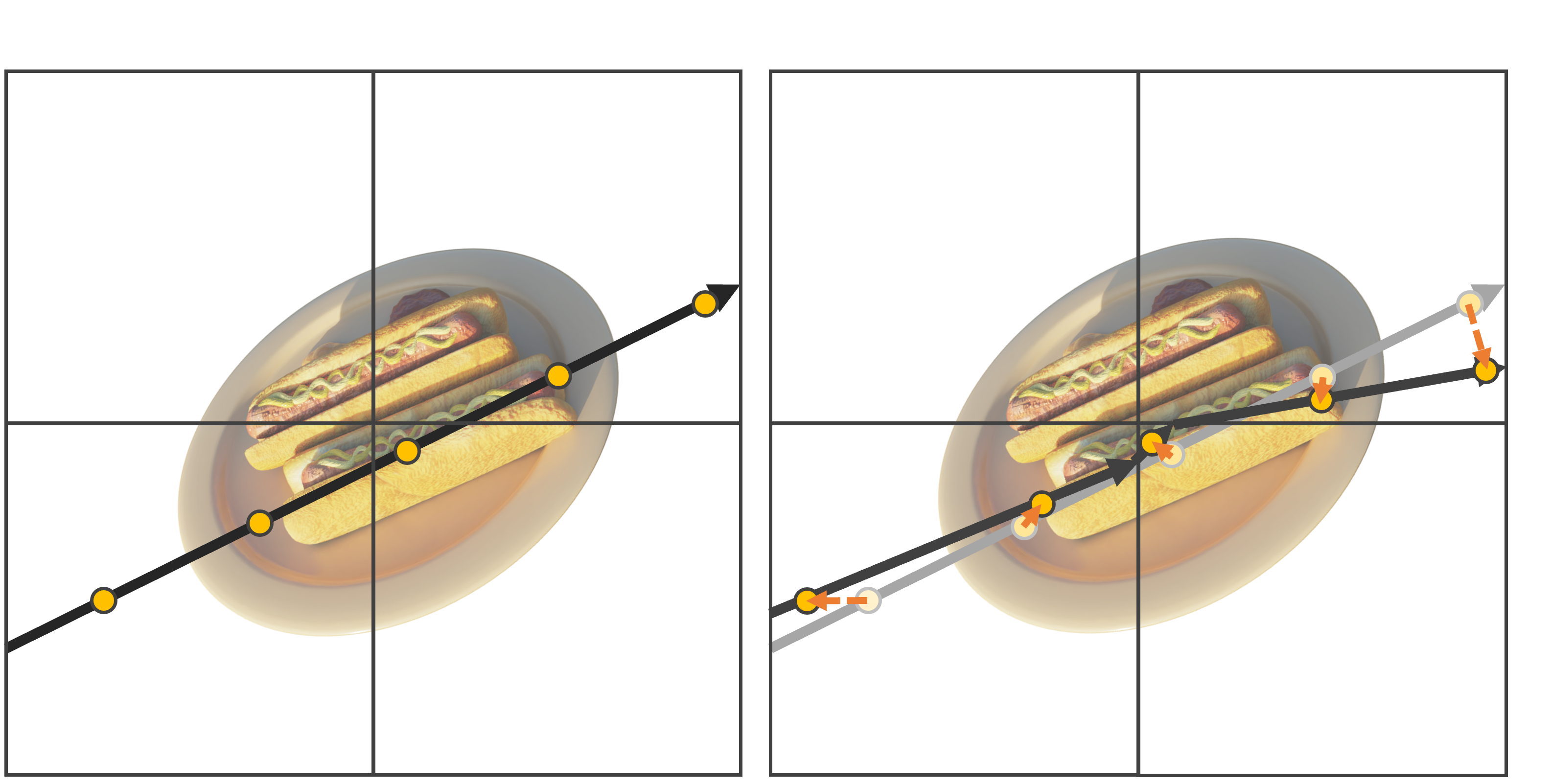}}
    \caption{Illustration about ``zigzag problem'' of the smooth interpolation.
    Left: ray casting with linear interpolation, Right: ray casting with smooth interpolation in \cref{eq:smooth_weighting}.
    As shown on the right, if we directly use smooth interpolation in a certain hypercube, the sampled points are getting closer to the edges.
    }
    \label{fig:zig}
    \end{minipage}
    \end{center}
\end{figure}

%% file: Tables/lambda.tex
\begin{table}[hbt!]
  \caption{Hyperparameter Search for the $\lambda$. 
  Experiments are conducted on the \textit{Hotdog} in the NeRF-Synthetic.
  }
  \label{tab:lambda}
  \vskip 0.15in
  \begin{center}
  \begin{small}
  \begin{sc}
  \begin{tabular}{ c  c  c c c  }
    \toprule
    \multirow{2}{*}{} 
    & \multirow{2}{*}{$\quad \lambda \quad $} 
    & \multicolumn{3}{c}{\textbf{Evaluation Metric}} 
    \\
    \cmidrule{3-5}
    & 
    & {Rotation $({}^{\circ})$ $\downarrow$} 
    & {Translation $\downarrow$} 
    & {PSNR $\uparrow$} 
    \\  
    \midrule

    (a)
    & 0.5
    & \text{0.433}
    & \text{2.003}
    & \text{32.14}
    \\

    (b)
    & 1
    & \text{0.234}
    & \text{1.124}
    & \text{35.41}
    \\

    (c)
    & 2
    & \text{0.229}
    & \text{1.113}
    & \text{35.48}
    \\

    (d)
    & 4
    & \text{0.241}
    & \text{1.128}
    & \text{35.15}
    \\
    
    \bottomrule
  \end{tabular}
\end{sc}
\end{small}
\end{center}
\vskip -0.1in
\end{table}

%% file: Appendix_B.tex
\section{Additional Experiments}
\label{sec:b}

\subsection{Re-implementation of Instant-NGP}
\label{sec:b.1.}
\subsubsection{Implementation Details}
We follow the implementation settings from the original Instant-NGP ~\cite{mueller2022instant} and BARF~\cite{lin2021barf} except for the number of layers (Refer to Section~\ref{sec:4.2.3}).
For the NeRF-Synthetic and the LLFF, we resize the images for each dataset to $400 \times 400$ and $480 \times 640$, respectively. 
We use the sinusoidal encoding with $4$ levels for view-direction encoding, employed by \citet{mildenhall2020nerf}.
For ray casting, we randomly sample $1024$ rays and $N=128$ samples per ray at each optimization step. 
Notice that the sum of all weights normalizes all the reparameterized weights to ensure that their sum equals one.
We use the Adam optimizer and train all models for 200K iterations, with a learning rate of $5\times 10^{-4}$ that exponentially decays to $1\times 10^{-4}$.
For multi-level learning rate scheduling, we set the scheduling interval $[t_s, t_e] = [20\text{K}, 100\text{K}]$, which is between 10$\%$ to 50$\%$ progress of the entire training. 

\subsubsection{Quantitative Results}
This section presents the performance of our re-implemented version of multi-resolution hash encodings. 
The following experiments were conducted with ground-truth camera poses to compare it with the original Instant-NGP~\cite{mueller2022instant}.
In Table~\ref{tab:ngp_synthetic} and Table~\ref{tab:ngp_llff}, we confirm that our implemented version gets similar results compared with the original. 
\input{Tables/ngp_synthetic.tex}

\input{Tables/ngp_llff.tex}

\subsection{Real World Scenes from COLMAP Initialization}
\label{sec:b.2.}
The manuscript investigates the challenging problem of jointly optimizing unknown camera poses and scene reconstruction using multi-resolution hash encoding. 
In this section, we conduct additional experiments on the LLFF dataset with the COLMAP~\cite{2016sfm} camera pose initialization. 
Given that the estimated poses from COLMAP are inaccurate, we further refine the provided poses through the joint optimization of camera poses and scene reconstructions.
\input{Tables/llff_colmap_appendix.tex}

Table~\ref{tab:llff_colmap} shows the quantitative results on the LLFF dataset with the COLMAP-initialized poses and the further pose refinement using the proposed method. 
Here, '\textit{w/} Pose Refinement' denotes the joint optimization of camera poses and scene reconstructions by the proposed method.

As shown in row (b) compared with (a), the results show that the proposed method achieves better view synthesis quality for some scenes, such as \textit{Fortress, Leaves, Room}.
This supports that pose registration (from scratch) can often align more accurate poses for scene reconstruction, even better than the classical geometry-based approaches~\cite{2016sfm}. 
Additionally, comparing rows (a) and (c) in the Table, it is observed that the pose registration from the COLMAP initialization significantly outperforms the COLMAP (fixed) or the from-scratch.
Informed by this analysis, since the entire pipeline is lighter than the other vanilla NeRF architectures, the proposed method is expected to be useful in real-world problems of novel-view synthesis with imperfect camera poses, even for quite accurate ones from the COLMAP.

\subsection{Qualitative Results}
\label{sec:b.4.}
In Figure~\ref{fig:qual}, we present the qualitative results of the proposed method in the fourth column.
For comparison, the ground-truth image and the rendering of two previous works, BARF~\cite{lin2021barf} and GARF~\cite{2022garf}, are also presented in the first, second, and third columns, respectively. 
The ground-truth is shown at the beginning of each row.
The qualitative results generally show that our method gives the finer-grained novel-view synthesis with the camera pose refinement.

Figure~\ref{fig:qual} is shown in the next page.

\input{Figures/qual.tex}

%% file: Tables/ngp_synthetic.tex
\begin{table*}[hbt!]
  \centering
  \small
  \vspace{-0.7em}
  \caption{
      Quantitative results of our re-implemented multi-resolution hash encoding (PSNR $\uparrow$) on the NeRF-Synthetic.
  }
  \label{tab:ngp_synthetic}
  \vskip 0.15in
  \begin{adjustbox}{width=0.9\textwidth} 
  \begin{tabular}{ c c c c c  c c c c  c }
    \toprule
    \multirow{2}{*}{Model} 
    & \multicolumn{9}{c}{NeRF-Synthetic}  \\ 
    \cmidrule(r){2-10}   
    & Chair 
    & Drum
    & Ficus
    & Hotdog 
    & Lego
    & Materials 
    & Mic 
    & Ship 
    & \textit{Average}
    \\
    \midrule 
    \citet{mueller2022instant}
    & \textbf{35.00}
    & 26.02
    & 33.51
    & \textbf{37.40}
    & \textbf{36.39}
    & \textbf{29.78}
    & \textbf{36.22}
    & \textbf{31.10}
    & \textbf{33.18}
    \\

    {Ours}
    & 34.91 
    & \textbf{26.13}
    & \textbf{33.52}
    & 37.19
    & 36.23
    & 29.69
    & 36.17
    & 31.08
    & 33.12
    \\
    \bottomrule
  \end{tabular}
  \end{adjustbox}
  \vskip -0.1in
\end{table*}

%% file: Tables/ngp_llff.tex
\begin{table*}[hbt!]
  \centering
  \small
  \caption{
      Quantitative results of our re-implemented multi-resolution hash encoding (PSNR $\uparrow$) on the LLFF.
  }
  \label{tab:ngp_llff}
  \vskip 0.15in
  \begin{adjustbox}{width=0.9\textwidth} 
  \begin{tabular}{ c c c c c  c c c c  c }
    \toprule
    \multirow{2}{*}{Model} 
    & \multicolumn{9}{c}{LLFF}  \\ 
    \cmidrule(r){2-10}   
    & Fern
    & Flower
    & Fortress
    & Horns
    & Leaves
    & Orchids
    & Room
    & T-Res
    & \textit{Average}
    \\
    \midrule 
    \citet{mueller2022instant}
    & \textbf{25.87}
    & 26.52
    & 27.96
    & \textbf{26.60}
    & \textbf{18.93}
    & \textbf{20.31}
    & \textbf{31.96}
    & 26.50
    & \textbf{25.58}
    \\
    
    {Ours}
    & 25.83
    & \textbf{26.56}
    & \textbf{28.00}
    & 26.46
    & 18.89
    & 20.15
    & \textbf{31.96}
    & \textbf{26.51}
    & 25.55
    \\
    
    \bottomrule
  \end{tabular}
  \end{adjustbox}
  \vskip -0.1in
\end{table*}

%% file: Tables/llff_colmap_appendix.tex
\begin{table*}[hbt!]
\renewcommand\thetable{4}
  \centering
  \small
  \vspace{-1em}
  \caption{
      Quantitative results of the proposed method in the LLFF dataset with the COLMAP initialization (PSNR $\uparrow$).
  }
  \label{tab:llff_colmap_appendix}
  \vskip 0.15in
  \begin{adjustbox}{width=1\textwidth} 
  \begin{tabular}{c  c c  c c c c c c c c  c }
    \toprule
    & \multicolumn{2}{c}{Experimental Setting}
    & \multicolumn{9}{c}{LLFF}  \\ 
    \cmidrule(r){2-3}   
    \cmidrule(r){4-12}   
    
    & \textit{w/} COLMAP 
    & \textit{w/} Pose Refinement
    & Fern
    & Flower
    & Fortress
    & Horns
    & Leaves
    & Orchids
    & Room
    & T-Res
    & \textit{Average}
    \\
    \midrule 
    (a)
    & \checkmark
    & 
    & \underbar{25.83}
    & \underbar{26.56}
    & 28.00
    & \underbar{26.46}
    & 18.89
    & \underbar{20.15}
    & 31.96
    & \underbar{26.51}
    & \underbar{25.55}
    \\
    (b)
    &
    & \checkmark
    & 24.62 
    & 25.19
    & \underbar{30.14}
    & 22.97
    & \underbar{19.45}
    & 20.02
    & \underbar{32.73}
    & 23.19
    & 24.79
    \\
    (c)
    & \checkmark
    & \checkmark
    & \textbf{26.41}
    & \textbf{28.00}
    & \textbf{30.99}
    & \textbf{27.35}
    & \textbf{19.97}
    & \textbf{21.26}
    & \textbf{33.02}
    & \textbf{26.83}
    & \textbf{26.73}
    \\
    
    \bottomrule
  \end{tabular}
  \end{adjustbox}
  \vskip -0.1in
\end{table*}

%% file: Figures/qual.tex
\begin{figure*}[hbt!]
    \vskip 0.2in
    \begin{center}
    \centerline{\includegraphics[width=0.9\textwidth]{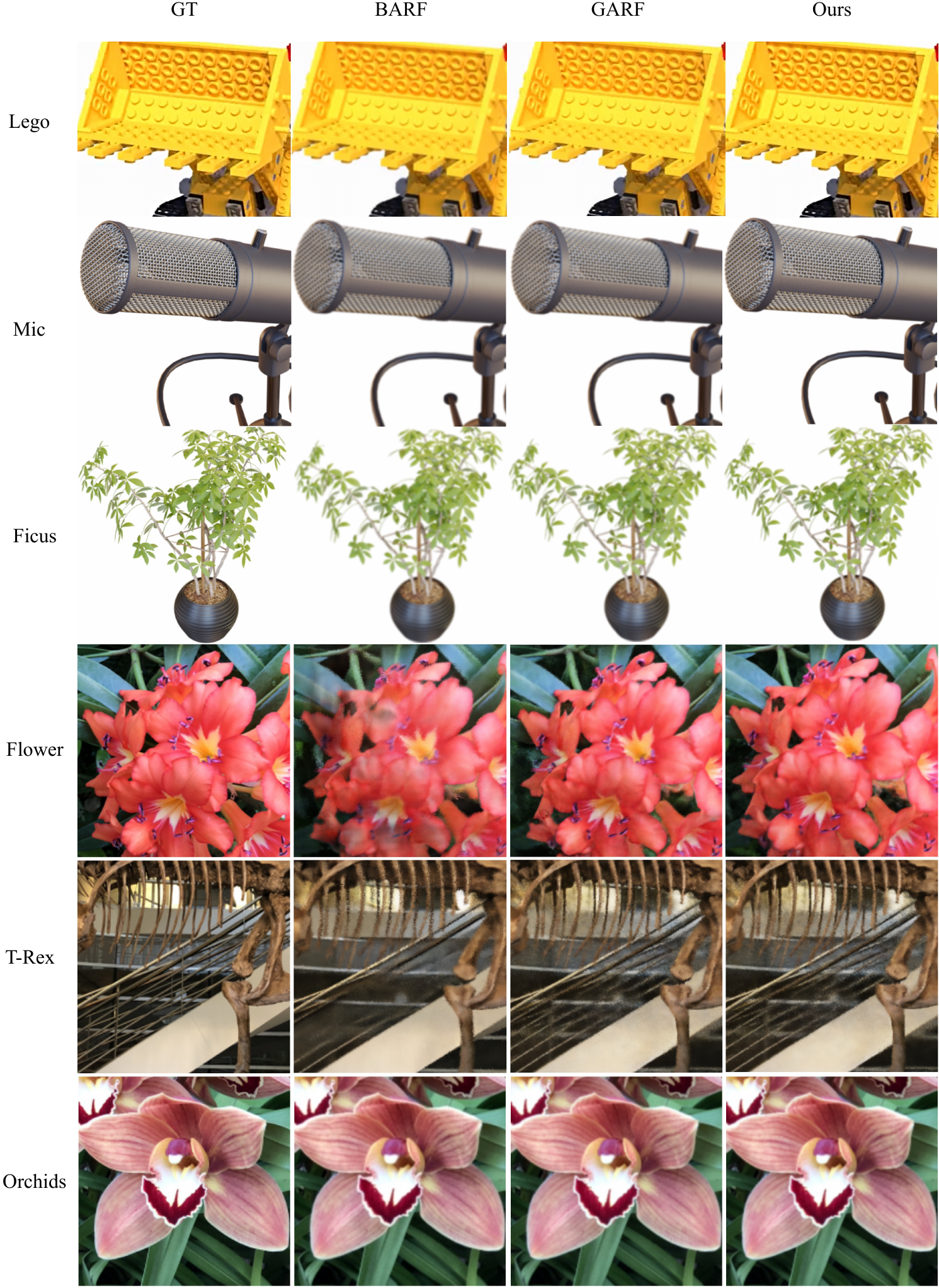}}
    \vskip -0.15em
    \caption{Qualitative results of the proposed method.
    }
    \label{fig:qual}
    \end{center}
    \vskip -0.2in
\end{figure*}